\pdfoutput=1
\documentclass[letterpaper, 10 pt, conference]{ieeeconf}
\IEEEoverridecommandlockouts
\usepackage{amsmath,amssymb}
\makeatletter\let\NAT@parse\undefined\makeatother
\usepackage[numbers,sort&compress]{natbib}
\usepackage{graphicx}
\usepackage{booktabs}
\usepackage{hyperref}
\hypersetup{colorlinks=true,linkcolor=[rgb]{0,0,0.55},citecolor=[rgb]{0,0.35,0},urlcolor=[rgb]{0,0,0.55}}
\hypersetup{pdftitle={Audit Before You Commit: Locating Belief Failures in Active Identification for One-Shot Manipulation},pdfauthor={Mohamed Abouagour, Byung-Cheol Min}}
\makeatletter
\def\ps@arxivfirst{\let\@mkboth\@gobbletwo
\def\@oddhead{\smash{\parbox[b]{\textwidth}{\centering\footnotesize This work has been submitted to the IEEE for possible publication. Copyright may be transferred without notice, after which this version may no longer be accessible.}}}%
\def\@oddfoot{\parbox[t]{\textwidth}{\centering\footnotesize The authors are with the SMART Lab, Luddy School of Informatics, Computing, and Engineering, Indiana University, Bloomington, IN, USA (e-mail: moabouag@iu.edu; minb@iu.edu).}}%
\let\@evenhead\@oddhead\let\@evenfoot\@oddfoot}
\makeatother

\newcommand{\anonrepo}{https://anonymous.4open.science/r/audit-before-you-commit/}
\usepackage{xspace}
\makeatletter\g@addto@macro\normalsize{\setlength\abovedisplayskip{3pt plus 1pt}\setlength\belowdisplayskip{3pt plus 1pt}}\makeatother
\newcommand{\Param}{\theta}

\newcommand{\miss}{\mathrm{miss}}

\newcommand{\BlindMean}{Blind-Bayes\xspace}
\newcommand{\BlindRob}{Blind-Robust\xspace}
\newcommand{\ProbeRob}[1]{Probe-Robust(#1)\xspace}
\usepackage[font=footnotesize]{caption}
\title{\LARGE\bf Audit Before You Commit: Locating Belief Failures in\\
Active Identification for One-Shot Manipulation}
\author{Mohamed Abouagour and Byung-Cheol Min}

\begin{document}
\maketitle
\thispagestyle{arxivfirst}
\pagestyle{empty}

\begin{abstract}
A robot that probes a few times before one irreversible action, such as tapping a surface before inserting a peg, must decide when the evidence is enough to commit. We argue that this decision rests on two conditions that existing methods do not separate: the belief must still cover the truth in the coordinate that decides the action, and the failure model that scores actions must track realized failure. We audit both conditions separately, offline and with ground truth, on a deployed probe-then-commit pipeline: a particle belief, a scenario failure score, and one commit. On simulated insertion, more taps sharpen the belief while the truth leaves its support on $16.9\%$ of episodes and the failure score turns optimistic by $0.31$. Conformal calibration restores coverage but not the decision: confidently wrong instances still pass a confidence gate. The audit's signatures instead point at the observation model, where a hand scan finds a $2.1$\,mm error in the tap boundary. Correcting that one number cuts failure from $0.354$ to $0.112$ on untouched instances and transfers unrefitted to a second engine, while in a third engine the same audit suggests an execution-model mismatch instead. Across seven task families in three engines, a few probes at a fixed executor reduce miss or failure. On a physical arm inserting a tool into a rigid pocket by touch, the gain and the audit's two conditions reproduce, and replaying the recorded taps under an injected model error shows the audit's signature on real data. Additional materials are available at \url{https://sites.google.com/view/auditbeforeyoucommit}.
\end{abstract}

\section{Introduction}
\begin{figure}[t]
\centering
\includegraphics[width=0.97\columnwidth]{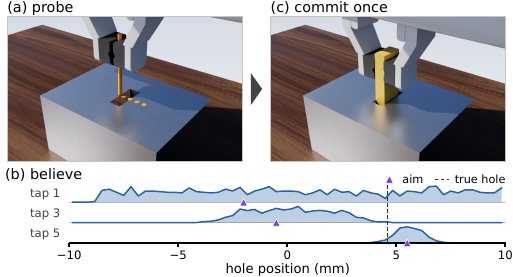}
\vspace{-2pt}
\caption{\textbf{Probe, believe, commit.} One simulated insertion episode: (a) surface-height probes; (b) peak-normalized particle beliefs after taps 1, 3, and 5, with the true hole position and candidate aims; (c) insertion at the fifth aim.}
\label{fig:overview}
\end{figure}

\begin{figure*}[t]
\centering
\includegraphics[width=0.95\textwidth]{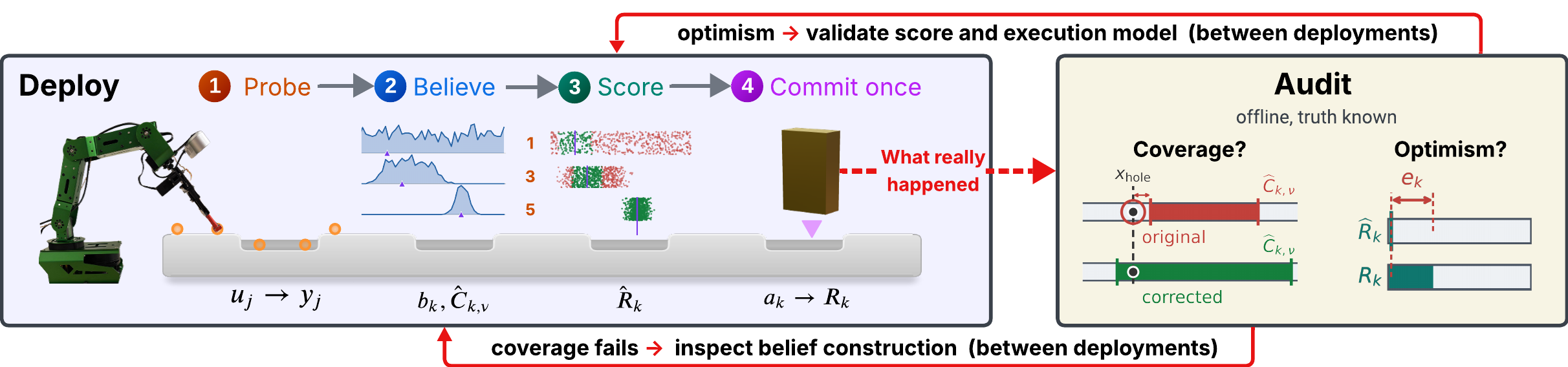}
\vspace{2pt}
\caption{\textbf{Probe-then-commit execution and offline audit.} Probes update the belief; retained scenarios select an action and failure score. Ground-truth audits assess decision-coordinate coverage and mean score optimism separately. Coverage failures prompt belief-model inspection; optimism prompts score and execution-model validation. Neither uniquely identifies the cause. Corrections apply between deployments, not within an episode.}
\label{fig:pipeline}
\vspace{-10pt}
\end{figure*}

Consider inserting a peg into a socket whose lateral position is known to a few millimetres. The robot may tap the surface first, but the insertion happens once (Fig.~\ref{fig:overview}); Fig.~\ref{fig:pipeline} shows the
pipeline and the audit this paper places on it. If the prior spans positions that call for incompatible aims, a $\theta$-independent policy must choose without instance-specific evidence; pushing a box of unknown friction poses the same problem. Domain randomization (DR)~\cite{tobin-2017-domain,peng-2018-dynamicsdr} can train such a prior-averaged policy; history-conditioned DR can instead infer dynamics online. We study the former, commit-before-feedback case, where a few exploratory interactions can supply what is missing.

In assembly, maintenance and remote manipulation a robot may contact a socket before inserting a fragile connector, or probe friction before moving an unrecoverable object: interactions too few for retraining, before a task that cannot be repeated. The problem this paper addresses is deciding when a few such probes provide enough evidence for the robot to commit safely to one irreversible action.

Existing work does not settle this. Active identification and learned exploration choose informative probes and switch policies on uncertainty~\cite{memmel-2024-asid,aoyama-2025-pokestrike,wang-2026-phys2real}; filter-consistency tests~\cite{barshalom-2001-estimation} and conformal prediction~\cite{angelopoulos-2023-conformal} check a belief against its own data or wrap it in a coverage guarantee. All of them judge the belief. A one-shot commitment depends on two separate conditions: that the belief still covers the truth in the coordinate that decides the action, and that the failure model the executor scores actions with tracks realized failure. A belief can meet the first and fail the second, or explain every tap it has seen while excluding the truth, and calibrating one condition does not repair the other.

We therefore audit both, separately and offline, on the deployed executor: the robot probes, updates a particle belief, and commits once through a scenario failure score. On instances with known ground truth, we measure decision-coordinate coverage and mean score optimism separately. Coverage calibration adjusts the retained set, while a validation-tuned execution margin addresses score optimism; neither certifies failure risk conditional on commitment. On simulated insertion, both audit conditions worsen as taps accumulate. Calibration improves coverage but not the decision. A hand scan identifies a $2.1$\,mm observation-boundary error in a model that had been checked against the simulator before any audit and had passed; the error lay in a band that check never sampled. Its correction cuts failure from $0.354$ to $0.112$ on untouched instances and transfers unrefitted to PyBullet. In PhysX, the audit instead suggests a remaining capture-model mismatch. The same pipeline runs on six further task families and on a physical arm, where a touch-probe run into a rigid pocket reproduces the gain, passes the audit, and, replayed under an injected model error, shows the audit's signature on real taps.


\textbf{Contributions.} (i)~\emph{A two-condition audit of the implemented executor}: decision-coordinate coverage, also across repeated episodes of one instance, and mean optimism of the modeled failure score, measured separately (Section~\ref{sec:calib}). (ii)~\emph{Particle-based coverage calibration, and its decision-level limit}: a score that continues past the particle hull, calibrated at fixed or simultaneous budgets and kept distinct from the performance of commit gates (Sections~\ref{sec:calib}, \ref{sec:analysis}). (iii)~\emph{Audit-guided model inspection with independent confirmation}: the scan, the frozen correction, its confirmation on untouched instances, its transfer across engines, and a physical-arm run whose recorded taps replay under injected faults (Section~\ref{sec:exp}).

\section{Related Work}
\noindent\textbf{DR and its limits.} DR trains policies over distributions of physical
parameters~\cite{tobin-2017-domain,peng-2018-dynamicsdr}; history-conditioned
variants infer dynamics from interaction history~\cite{kumar-2021-rma}. Theory bounds horizon-dependent sim-to-real gaps~\cite{chen-2022-understanding-dr,wagenmaker-2024-sim2real} and favours smooth linear-quadratic control~\cite{mania-2019-certainty,fujinami-2025-lqrdr}.

\noindent\textbf{Active identification.} Dual control~\cite{feldbaum-1960-dual,barshalom-1974-dual} selects actions for information and task value at once; ASID~\cite{memmel-2024-asid}, Poke-and-Strike~\cite{aoyama-2025-pokestrike} and ActivePusher~\cite{zhong-2026-activepusher} learn informative interactions, interactive perception chooses actions for information gain~\cite{kruzliak-2024-interactive,bohg-2017-interactive}, Phys2Real~\cite{wang-2026-phys2real} conditions on probed estimates with their uncertainty, and identification can be done offline~\cite{ramos-2019-bayessim,tiboni-2023-dropo}.

\noindent\textbf{Distribution-free calibration.} Split conformal prediction turns a score into a marginal coverage guarantee under exchangeability~\cite{lei-2018-distribution-free,angelopoulos-2023-conformal}. Robotics uses calibrated sets in planning~\cite{lindemann-2023-conformal,dixit-2023-adaptive-conformal} and particle-based conformal sets for contact-aware motion prediction in tight-tolerance insertion~\cite{capture-2026}. That work calibrates future configurations; ours calibrates the decision coordinate of a latent instance and certifies coverage only (Section~\ref{sec:calib}).

\noindent\textbf{Model criticism.} Filter consistency tests~\cite{barshalom-2001-estimation} and Bayesian model criticism~\cite{gelman-1996-ppc,dawid-1984-prequential} ask whether a model explains what it has seen, a condition that passes while both of ours fail (Section~\ref{sec:analysis}). Since information has no decision value when the optimal action is insensitive to the unknown~\cite{howard-1966-voi}, we hold the probe selector fixed and audit the belief.

\section{The Probe-Then-Commit Pipeline}\label{sec:method}
An episode draws a latent instance parameter $\Param^\star$ (such as
friction, mass, or hole pose) from an environment distribution
$P_{\rm env}$, and the robot's belief starts at a prior $b_0=p_0$. At step
$j$, an exploratory policy selects probe $u_j$ from the preceding probe history,
observes a noisy outcome $y_j$, and updates belief $b_j$. After $k$ probes,
the robot commits to one action $a$ with task loss
$\miss(a,\Param^\star)\ge0$.

For a fixed observed state and target, let $\rho$ be a risk functional and
define the risk-matched no-probe and active actions as
\begin{equation}
\begin{aligned}
a_0^\rho&\in\arg\min\nolimits_a\rho_{p_0}[\miss(a,\Param)],\\
a_k^\rho&\in\arg\min\nolimits_a\rho_{b_k}[\miss(a,\Param)],
\end{aligned}
\label{eq:riskmatched}
\end{equation}
the risk taken over $\Param\sim p_0$ and $\Param\sim b_k$. Two axes separate a no-probe baseline from our pipeline, information and risk functional. \BlindMean takes no probes and executes under the expected value of the loss
the task reports. We compute the prior-averaged optimum directly, so the baseline is a converged DR policy, not an undertrained one. \BlindRob takes no probes but executes under the
set-based objective below, and \ProbeRob{k} takes $k$ probes under that same
objective.

\noindent\textbf{Belief update.} We maintain $b_k(\Param)\propto
p_0(\Param)\prod_{j=1}^k\widetilde p_f(y_j\mid u_j,\Param)$ with $N=1024$
particles ($512$ for SAPIEN shuffleboard), each probe reweighting particle $i$
by $\widetilde p_f(y_j\mid u_j,\Param_i)\propto
\exp(-\|y_j-f(u_j,\Param_i)\|^2/2\sigma^2)$, where $f$ is an analytic forward
model and $\sigma$ matches each testbed's configured observation noise. The filter is a particle approximation of $b_k$: when the effective sample size falls below $N/2$ we resample systematically (one uniform offset, $N$ equally spaced points on the cumulative weight), jitter each copy, clip to the prior box and reset the weights to $1/N$. Execution and calibration read the same weights after any resampling at that step, sort them descending with ties in particle-index order, and retain the shortest prefix reaching mass $\nu$, so retained subsets are nested in $\nu$. Directly after a resampling all weights tie, and the prefix is the first $\lceil\nu N\rceil$ copies in parent order, an arbitrary fifth of the cloud dropped rather than its lightest members.

\noindent\textbf{The insertion model.} For insertion $\Param=(x_{\rm hole},c)$, hole position $x_{\rm hole}\sim U[-10,10]$\,mm and clearance scale $c\sim U[0.7,1.3]$; the slot half-gap is $\mathrm{hg}(c)=4+0.6c$\,mm, the probe tip radius $r=1.2$\,mm, and the jitter is $\mathcal N(0,0.6\,\mathrm{mm})$ in $x_{\rm hole}$ and $\mathcal N(0,0.006)$ in $c$. A tap at $x_t$ reads the plate top or the pocket floor $15$\,mm below it, with $\sigma=0.5$\,mm noise. The model as deployed, and as corrected in Section~\ref{sec:exp}, reads pocket when
\begin{equation}
|x_t-x_{\rm hole}|<\mathrm{hg}(c)-r\ \text{(original)},\ \ \mathrm{hg}(c)+\delta\ \text{(corrected)},
\label{eq:tap}
\end{equation}
with $\delta=0.905$\,mm fitted on the development instances: the correction moves this one boundary by $2.1$\,mm and nothing else. The commit action is an aim $a$ on a $0.5$\,mm grid over $[-10,10]$\,mm. The modeled loss is binary, failure when $|a-x_{\rm hole}|\ge w(c)=0.6c+1.1$\,mm, a capture half-width calibrated against the simulator before any audit; the simulator outcome is separate: the peg is driven down at $a$ and the episode fails if it reaches less than $60\%$ of the pocket depth. The original boundary was checked against the simulator on the nine tap positions before any audit; the only injected faults are the five of Table~\ref{tab:faults}.

\noindent\textbf{Posterior-aware execution.} Let $C_{k,\nu}$ be a measurable credible
region with posterior mass at least $\nu$. The primary objective is
\begin{equation}
a^{C}_{k,\nu}=\arg\min_a\;\max_{\Param\in C_{k,\nu}}\;\miss(a,\Param).
\label{eq:pess}
\end{equation}
The implementation replaces $C_{k,\nu}$ with the top-weight $\nu$-mass
particle subset $\widehat C_{k,\nu}$, a scenario surrogate. Continuous-loss
tasks minimize the maximum miss over it, a rule we call \emph{scenario
minimax}. For binary grasp and insertion a literal maximum is one for almost
every action, so we minimize the \emph{scenario failure fraction}, the
unweighted fraction of retained particles that fail. This fraction is a decision score, not a posterior failure probability or a certified bound, which is why insertion audits and calibrates it against realized failure and never invokes the continuous-loss minimax bound. Posterior risk minimization~\cite{wu-2018-bro} and posterior $\mathrm{CVaR}_\lambda$~\cite{rockafellar2000optimization,rajeswaran-2017-epopt,rigter-2021-rabamdp} are the loss-space alternatives to \eqref{eq:pess}; neither unimodality nor $\nu=\lambda$ makes them equivalent. The suite fixes $\nu=0.8$.

\noindent\textbf{Probe protocol and cost.} Within an episode $\Param^\star$ is fixed
and the task state is restored after each probe: the probe retracts to its
start pose above the plate before the peg descends; a pushed object is
returned to its start pose; each dart or hinge probe is an independent
launch. Probe duration is not in the task loss. Pricing it as $J_\lambda(k)=\mathrm{miss}(k)+\lambda k$ on the frozen curves of one fixed executor (\BlindRob at $k=0$, \ProbeRob{k} at $k\in\{2,5\}$) gives two prices per family. Above $\lambda=1.2$\,cm per probe on darts, $1.7$\,cm on 2-D pushing, $0.17$\,rad on the hinge and $35$\,cm on 1-D pushing a blind commit beats every budget; below $0.06$\,cm, $1.7$\,cm, $0.012$\,rad and $1.9$\,cm five probes are the optimum, and between the two prices two probes are.

\section{Offline Audit and Calibration}\label{sec:calib}
For the ideal continuous-loss minimax objective, a correct loss model and a $C_{k,\nu}$ containing the full true parameter give the robot a bound. Coverage of a projected particle interval establishes neither, and insertion uses a binary scenario score besides. We therefore audit decision-coordinate coverage and mean agreement between modeled and realized failure separately.

\noindent\textbf{The audit.} We measure coverage in a decision coordinate $z(\Param)$, the hole position for insertion. \emph{Episode coverage} is the probability that this coordinate lies in the interval spanned by the retained particles. \emph{Cluster coverage} requires coverage in all $R$ repeats of the same instance. \emph{Support escape} means that the coordinate lies outside the range of all current particles. For insertion, $R_k$ and $\widehat R_k$ denote the mean realized binary failure and mean modeled scenario score on the evaluated episodes, and $e_k=R_k-\widehat R_k$ measures the robot's optimism. The audit is offline: it needs the ground truth of frozen test instances, and its corrections apply between deployments, never inside the one-shot episode.

\noindent\textbf{The episode score.} Sort the particles of repeat $\ell$ of instance
$i$ after $k$ probes by weight, write $W_m$ for the cumulative mass of the
first $m$, $I_m$ for the range of the decision coordinate $z$ over them, and
$m(t)=\min\{m:W_m\ge t\}$ for the prefix the executor retains at threshold
$t$. The score is the mass of the shortest prefix covering the truth,
$s_{i,\ell,k}=\min\{W_m:z_i^\star\in I_m\}$, so that
$s_{i,\ell,k}\le t\Rightarrow z_i^\star\in I_{m(t)}$. It continues past $1$ when the truth escapes the hull, $s=1+d(z_i^\star,I_{\rm hull})/D$ for $D=20$\,mm the width of the decision domain; clipping at $1$ would hide the $5.5$ to $16.9\%$ of episodes that escape. For $t\le1$ the set is $I_{m(t)}$ and $s\le t$ implies coverage. The one exception is the plateau that zero-weight particles, last in the ordering, form at $W=1$: there the prefix for $t=1$ can stop short of the particle spanning the truth. Every coverage we report is therefore measured on the set as constructed. For $t>1$ the set is the hull grown by $(t-1)D$ on each side. It is represented by all $N$ particles plus nine points evenly spaced across the grown interval, each at the particles' median clearance scale, and an aim is scored by the unweighted failure fraction over these $N+9$ scenarios. A threshold above $1$ is an expansion distance, not a mass; on the cover schedule (the fixed five-tap sequence of Section~\ref{sec:exp}) $\widehat t_5=0.42$, so the expansion never engages.

\noindent\textbf{Instance-level calibration.} At a fixed budget $k$, instance $i$ contributes $s^{\rm rep}_{i,k}=\max_{\ell\le R}s_{i,\ell,k}$. For a pre-specified budget set $\mathcal K$, use $S_i=\max_{k\in\mathcal K}s^{\rm rep}_{i,k}$ to obtain simultaneous coverage across budgets. With a fixed score construction and exchangeable calibration and test clusters, the $q$-th smallest calibration score, $q=\lceil(1-\alpha)(M_{\rm cal}+1)\rceil$, defines $\widehat t_k$ or the shared threshold $\widehat t$. The threshold is $+\infty$ if $q=M_{\rm cal}+1$. The resulting guarantee is marginal coverage of all $R$ repeats of a fresh cluster, and of all $k\in\mathcal K$ for the simultaneous construction, with probability at least $1-\alpha$. We use $1-\alpha=0.80$ and $R=4$.

\noindent\textbf{The margin.} Set coverage alone is insufficient, the capture-window model being optimistic too, so the executor scores each aim with the window shrunk to $\gamma\,w(c)$. The dimensionless $\gamma$ is searched over $\{1.00,0.98,\dots,0.30\}$ and set to the largest value whose mean predicted failure, over all calibration episodes and budgets, is at least the realized failure of the aim it selects. $\gamma$ changes action scoring and selection, but not the conformal score or retained set; it is $0.74$ on the cover schedule and $0.62$ under random probes. That margin is \emph{validation-tuned}, not distribution-free. The distribution-free alternative, a conformal $0.8$-quantile of each calibration instance's worst-budget optimism $\max_k e_{i,k}$, is $0.90$ on the cover schedule.

\noindent\textbf{The limit of history-based gates.} For a fixed probe policy let $H=(u_{1:k},y_{1:k})$ be the probe history, $g(H)\in\{0,1\}$ decide whether the robot commits, and $P^H_\theta$ be the law of the history under instance $\theta$. For instances $\theta^\star,\tilde\theta$ and any gate measurable with respect to $H$,
$\bigl|\Pr_{\theta^\star}[g(H)=1]-\Pr_{\tilde\theta}[g(H)=1]\bigr|
\le\mathrm{TV}\bigl(P^H_{\theta^\star},P^H_{\tilde\theta}\bigr)$, the definition of total variation applied to the event $\{g(H)=1\}$. For a mixture of two instances with identical history laws, prior weights $\pi_1,\pi_2$ and disjoint action-success sets, any history-based rule with positive commitment probability fails among committed episodes at rate at least $\min\{\pi_1,\pi_2\}/(\pi_1+\pi_2)$. Correcting the observation model changes inference for a fixed history, whereas changing the probe policy can change the history laws. Our experiments therefore establish limits of the evaluated gates, not impossibility for every history-based gate; the hindsight oracle of Section~\ref{sec:analysis} uses realized outcomes and is outside this class.

\section{Experiments}\label{sec:exp}
\subsection{Testbeds and comparisons}
\begin{figure}[t]
\centering
\vspace{5pt}
\includegraphics[width=0.92\columnwidth]{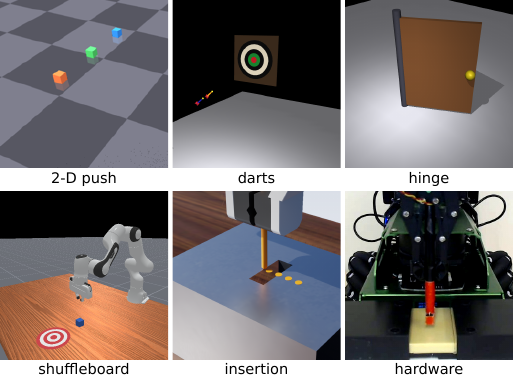}
\vspace{-3pt}
\caption{The simulated scenes, with their engines in Table~\ref{tab:setup}, and the hardware rig from the front camera: the probe over the machined pocket. The 1-D push is closed form and grasp is a
force-balance model with a MuJoCo lift check, so neither has a scene.}
\label{fig:testbeds}
\end{figure}

\noindent\textbf{Setup.} Table~\ref{tab:setup} lists the seven task families, their engines, latents, probes, actions, noise, budgets, instance counts and metrics, plus 2-D pushing ported to PyBullet as a transfer case; Fig.~\ref{fig:testbeds} shows the simulated scenes. The robot's probes are uniform random unless stated. A reproducibility script checks $788$ numerical invariants in the frozen result files. Pushing, darts and hinge use fixed truth and target sets ($3\times6$, $6\times3$, $6\times3$) with random seeds. Unless otherwise stated, confidence intervals use a cluster bootstrap keeping dependent episodes together, over truth--target instances with their seeds, or truth instances with their four repeats, the AUC intervals included. Calibration and test instances are independent draws from one generator with scores aggregated per instance, the exchangeability the conformal step assumes.

\begin{table}[h]
\vspace{5pt}
\caption{\textbf{Experimental setup.} Seven task families; simulation engines are MuJoCo, SAPIEN (PhysX), and PyBullet. $\sigma$ denotes probe-noise standard deviation; $n$ gives instances $\times$ seeds or insertion repeats, per method and budget. The 2-D pushing transfer uses PyBullet ($k=5$, $18{\times}6$).}
\label{tab:setup}
\centering\scriptsize
\setlength{\tabcolsep}{1.0pt}
\begin{tabular}{@{}lllllll@{}}
\toprule
family & engine & latents & probe / action & $\sigma$ & budgets $k$ & $n$; metric \\
\midrule
insertion & all three & $x_{\rm hole},c$ & tap / aim & $0.5$\,mm & 1--8, cover & $96{\times}4$; failure \\
1-D push & analytic & $\mu,m,k_d$ & pulse / impulse & $5$\,cm & 0, 2, 5 & $81{\times}6$; miss \\
2-D push & MuJoCo & $\mu,m,k_d$ & pulse / impulse & $5$\,cm & 0, 2, 5 & $18{\times}6$; miss \\
darts & MuJoCo & $k_d,m,g$ & throw / launch & $2$\,cm & 0, 2, 5 & $18{\times}8$; miss \\
hinge & MuJoCo & $f,b,m$ & coast / torque & $0.02$\,rad & 0, 2, 5 & $18{\times}16$; miss \\
shuffleboard & SAPIEN & $\mu,m,k_d$ & strike / strike & $5$\,cm & 0, 2, 5 & $18{\times}6$; miss \\
grasp & MuJoCo & $\mu,m$ & squeeze / grip & $2$\,mm & 0, 2, 5, 8 & $16{\times}80$; success \\
\bottomrule
\end{tabular}
\end{table}

\emph{Instance sets.} Every insertion set is drawn from the same prior box, none overlap, four repeats each. \emph{Development}: $24$ uniform draws, used for the schedule, the boundary scan and the robust-filter hyperparameters. \emph{Split~A}: $96$ Latin-hypercube instances, the held-out cover-schedule evaluation (Table~\ref{tab:insertion}). \emph{Split~B}: $96$ calibration and $96$ test instances per protocol, cover or random; calibration sets the level, the margin and the thresholds, test carries everything else. \emph{Test~C}: $96$ per protocol, drawn after the correction froze. The model is the original boundary of \eqref{eq:tap} unless marked corrected; Tables~\ref{tab:audit} and~\ref{tab:insertion} use fixed-budget levels, Section~\ref{sec:analysis} the budget-simultaneous threshold. Random $k{=}8$ support escape is $0.169$ on the Split~B audit run (Table~\ref{tab:audit}), $0.174$ on a second random-probe realization of the same instances in the fault harness (Table~\ref{tab:faults}, Fig.~\ref{fig:signatures}) and $0.164$ on Test~C (Table~\ref{tab:testc}); cover-schedule failure of the original model is $0.234$, $0.255$ and $0.326$ on A, B and C: three instance sets, not three procedures.

Insertion is the primary testbed. The unknown hole position spans $6\times$ the capture window, so a robot aiming blind fails on most instances while the oracle never fails. Taps are chosen from nine positions, $-12$ to $12$\,mm in $3$\,mm steps. The cover schedule is the one of the $\binom{9}{5}=126$ five-position subsets, each executed in increasing position order, with the lowest mean miss on the $24$ development instances, $[-12,-6,0,6,9]$\,mm, frozen before any held-out outcome was examined; Split~A of Table~\ref{tab:insertion} is its held-out evaluation. The same pipeline, with task-specific models, runs on the six other families and on cross-engine 2-D pushing; Table~\ref{tab:breadth} collects every family.

\begin{table}[h]
\vspace{5pt}
\caption{\textbf{Every family, blind against probed}, uncalibrated, five probes (two for shuffleboard, eight for grasp). Mean miss (m; rad for the hinge; mean\,/\,P90 for shuffleboard), failure for insertion on Split~A, success for grasp on its feasible subset; MJ and PB are MuJoCo and PyBullet. From \BlindMean to \BlindRob only the risk functional changes, from \BlindRob to \ProbeRob{k} only the information. Last column: paired reduction against the matched blind baseline, cluster-bootstrap interval. PyBullet uses the MuJoCo-fitted analytic model; a residual model reaches $0.115$.}
\label{tab:breadth}
\centering\scriptsize
\setlength{\tabcolsep}{1.3pt}
\begin{tabular}{@{}lcccl@{}}
\toprule
family & \BlindMean & \BlindRob & \ProbeRob{$k$} & paired $\Delta$ [95\%] \\
\midrule
insertion, A & $0.820$ & -- & $0.234$ & $0.586\,[0.492,0.674]$ \\
1-D push & $0.491$ & -- & $0.062$ & -- \\
2-D push, MJ & $0.252$ & $0.307$ & $0.221$ & $0.031\,[0.007,0.054]$ \\
2-D push, PB & $0.243$ & -- & $0.141$ & -- \\
darts & $0.026$ & $0.035$ & $0.010$ & $0.016\,[0.009,0.024]$ \\
hinge & $0.334$ & -- & $0.031$ & $-91\%$ \\
shuffleboard & $0.083$/$0.199$ & -- & $0.050$/$0.123$ & -- \\
grasp & $0.333$ & -- & $0.692$ & -- \\
\bottomrule
\end{tabular}
\end{table}

\noindent\textbf{Comparisons.} Table~\ref{tab:breadth} separates the two axes one at a time; the three paired reductions all exclude zero. Grasp is reported descriptively: three of six feasible truths improve by $0.36$ to $0.96$, three do not move, and with three ties an exact sign-flip test cannot fall below its $p=0.25$. On MuJoCo pushing a GRU trained on expected miss with the pipeline's schedule, prior, targets and action grid matches the pipeline under that objective: $0.153$ against $0.157$\,m over $15$ training seeds, difference $-0.004$\,m, $95\%$ CI $[-0.008,+0.000]$, inside a $\pm0.020$\,m equivalence margin fixed before the seeds were run. The release (\url{\anonrepo}) has the code, frozen configurations, seeds, and the verification script mapping every reported value to its file.

\subsection{The insertion audit}
In insertion, episode coverage of the original model's region falls from $0.94$ at $k=1$ to $0.77$ at $k=8$, below the nominal $0.80$ (Table~\ref{tab:audit}). Over the same budgets support escape rises from $5.5\%$ to $16.9\%$, and $e_8$ reaches $+0.31$. With calibration, cluster coverage sits within a few points of the $0.80$ target at every budget and mean optimism falls from $0.216$ to $0.048$ (Table~\ref{tab:audit}, calibrated columns).

\begin{table}[h]
\vspace{3pt}
\caption{\textbf{Insertion audit.} Split~B, random probes. Cluster coverage requires all $R=4$ repeats to be covered; $e_k=R_k-\widehat R_k>0$ indicates optimism. The bottom row averages $|\mathrm{coverage}-0.80|$ and $|e_k|$ over budgets.}
\label{tab:audit}
\centering\scriptsize
\begin{tabular}{@{}lcccc@{}}
\toprule
& \multicolumn{2}{c}{Cluster coverage} & \multicolumn{2}{c}{$e_k$} \\
\cmidrule(lr){2-3}\cmidrule(lr){4-5}
$k$ & original & calibrated & original & calibrated \\
\midrule
1 & $0.79$ & $0.81$ & $+0.115$ & $\mathbf{-0.003}$ \\
2 & $0.69$ & $0.82$ & $+0.176$ & $+0.022$ \\
3 & $0.55$ & $0.78$ & $+0.218$ & $+0.078$ \\
5 & $0.42$ & $0.76$ & $+0.263$ & $+0.047$ \\
8 & $0.46$ & $0.77$ & $+0.306$ & $+0.091$ \\
\midrule
mean $|\cdot|$ & \multicolumn{2}{c}{$0.219\to0.025$} &
\multicolumn{2}{c}{$0.216\to0.048$} \\
\bottomrule
\end{tabular}
\end{table}

\noindent\textbf{Cost of calibration.} On Split~B ($n=384$) the robot fails on $87.0\%$ of episodes without probes, $25.5\%$ with five probes, $26.0\%$ with set calibration and $30.2\%$ with set calibration and margin (Table~\ref{tab:insertion}). The paired failure increase for the full adjustment is $+0.047\,[+0.005,+0.091]$; set calibration contributes $+0.005\,[-0.010,+0.021]$, and adding the margin contributes $+0.042\,[0.000,+0.083]$. Over $50$ random re-splits of one $192$-instance pool the calibrated level varies ($0.75\pm0.25$, SD across splits) while failure ($0.328\pm0.025$) and held-out cluster coverage ($0.857$) do not; the margin stays optimistic in $64\%$ of splits: tuned, not guaranteed.

\begin{table}[h]
\centering
\vspace{5pt}
\caption{\textbf{Insertion failure by split and calibration setting.} Each row uses $n=384$ episodes; probed rows use the five-tap cover schedule. Instance sets are defined in Section~\ref{sec:exp}. Brackets show truth-cluster bootstrap intervals.}
\label{tab:insertion}
\scriptsize
\setlength{\tabcolsep}{4pt}
\begin{tabular}{@{}llccl@{}}
\toprule
Split & Method & Set cal. & Margin & Failure \\
\midrule
A & no probe & -- & -- & $0.820\,[0.779,0.859]$ \\
A & cover, $k{=}5$ & no & no & $0.234\,[0.159,0.312]$ \\
\midrule
B & \BlindRob & -- & -- & $0.870\,[0.836,0.901]$ \\
B & cover, $k{=}5$ & no & no & $0.255\,[0.211,0.299]$ \\
B & cover, $k{=}5$ & yes & no & $0.260\,[0.182,0.341]$ \\
B & cover, $k{=}5$ & yes & yes & $0.302\,[0.221,0.385]$ \\
\bottomrule
\end{tabular}
\end{table}

\subsection{Audit-guided correction and cross-engine transfer}
\begin{figure}[t]
\centering
\includegraphics[width=0.95\columnwidth]{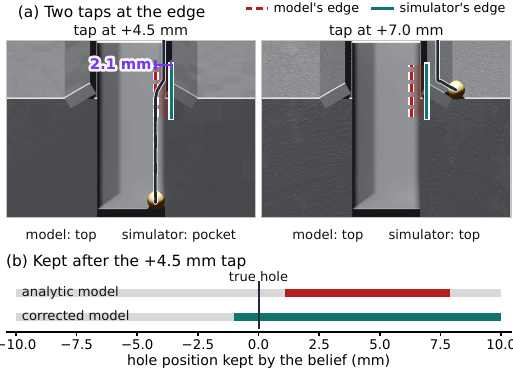}
\vspace{-3pt}
\caption{\textbf{The failure the audit found.} (a) Two taps as scale-accurate sections with the probe's simulated path. The analytic model reads pocket only inside the dashed edge; the simulator's finger slides down the chamfer and reads pocket out to the solid one, $2.1$\,mm further. (b) After the $+4.5$\,mm reading the analytic model's hypotheses exclude the true hole; the corrected model's keep it.}
\label{fig:teaser}
\end{figure}

\noindent\textbf{Audit-guided correction.} The audit does not name a model component by itself. Support escape, low late-stage surprise (Section~\ref{sec:analysis}) and optimistic execution scores together indicated a self-consistent misspecification of the observation model. We therefore scanned the tap model's top-or-pocket reading against lateral offset by hand, at $0.25$\,mm steps over $\pm7$\,mm and then by bisection on each side of the hole, on the $24$ development truths only. It found the analytic boundary $2.1$\,mm inside the simulator's, the $\delta$ of \eqref{eq:tap} at $+0.905$\,mm against the model's $-1.2$, identical on all $48$ scans, in a chamfer band missed by the original nine-position model check (Fig.~\ref{fig:teaser}). A tap there reads pocket while the
truth's particles predict top at $30\sigma$, so resampling removes the truth from the robot's belief.
Correcting that one boundary, everything else unchanged, removes the failure mode on the development instances, which motivated the correction and so cannot confirm it.

Table~\ref{tab:testc} lists every paired difference on instances drawn after the correction was frozen;
the same pattern holds, failure falling by $0.242$ at random $k=8$ and $0.167$ on the cover schedule. The original model reproduces the discovery signature on these instances, so it was not an artifact of the development set.
On the corrected belief the level, set on the frozen calibration split and applied to Test~C, prunes instead of expanding. The budget-simultaneous threshold is $0.046$ on the cover schedule and $0.085$ under random probes, against $1.004$ and $1.018$ for the original model. No Test~C episode escapes at the final budget of either protocol, so the committed pool holds none. Commitment is more reliable, not safe: failure at random $k=8$ remains $0.112$, an unconditional rate, not failure conditional on commitment, which we do not certify.

\begin{table}[h]
\caption{\textbf{Confirmation on untouched instances (Test C).} Drawn after the correction froze ($96$ per protocol, four repeats); both models see identical taps, differences paired and cluster-bootstrapped.}
\vspace{-3pt}
\label{tab:testc}
\centering\scriptsize
\setlength{\tabcolsep}{2.5pt}
\begin{tabular}{@{}lccl@{}}
\toprule
 & old & corr. & old $-$ corr.\ [95\%] \\
\midrule
escape, random $k{=}8$ & $0.164$ & $0.000$ & $+0.164\,[+0.125,+0.203]$ \\
episode cov., random $k{=}8$ & $0.768$ & $1.000$ & $-0.232\,[-0.279,-0.188]$ \\
optimism $e_8$, random & $+0.274$ & $-0.015$ & $+0.289\,[+0.234,+0.344]$ \\
failure, random $k{=}8$ & $0.354$ & $0.112$ & $+0.242\,[+0.188,+0.297]$ \\
failure, cover $k{=}5$ & $0.326$ & $0.159$ & $+0.167\,[+0.091,+0.247]$ \\
failure, cover $k{=}1$ & $0.721$ & $0.703$ & $+0.018\,[-0.016,+0.052]$ \\
cover $k{=}5$, set+margin & $0.294$ & $0.193$ & $+0.102\,[+0.031,+0.180]$ \\
\bottomrule
\end{tabular}
\end{table}

\noindent\textbf{Across engines.} The insertion study above is entirely in MuJoCo, so we rebuilt the scene parameter for parameter in PyBullet and in PhysX (SAPIEN) and ran the same belief, protocol and audit against two solvers we never tuned. With $\mathrm{hg}$ the slot half-gap of \eqref{eq:tap}, the pocket-reading boundary sits at $\mathrm{hg}+0.929$\,mm in PyBullet and $\mathrm{hg}-0.209$\,mm in PhysX against MuJoCo's $+0.905$ and the analytic model's $-1.2$. The capture window is $1.21\times$ the analytic one in PyBullet, with SD $0.69$ of that ratio across the $24$ development truths, and $0.55\times$ in PhysX.
In PyBullet the audit repeats MuJoCo's verdict (Table~\ref{tab:engines}): escape climbs with the budget, from $0.055$ at $k=1$ to $0.167$ at $k=8$ (MuJoCo $0.174$), coverage falls below nominal, and the executor turns optimistic. The MuJoCo-fitted offset, \emph{not refitted}, removes the escape and the optimism and cuts failure from $0.307$ to $0.109$, paired $-0.198\,[-0.253,-0.143]$; PyBullet's own scan buys no more ($-0.200\,[-0.255,-0.148]$). In PhysX the verdict changes. The analytic boundary is only $1.0$\,mm inside PhysX's, so coverage stays above the $0.80$ nominal (cluster coverage $0.72$) while the executor is optimistic by $+0.369$ and fails on $0.474$. This is consistent with the error lying in the capture model rather than the tap model, an attribution we did not verify separately: the MuJoCo offset overshoots and does not help ($+0.060\,[-0.003,0.122]$), and PhysX's own offset removes every escape yet leaves optimism and failure where they were ($-0.005\,[-0.076,0.062]$). The optimism is not its to fix.

\begin{table}[h]
\vspace{5pt}
\caption{\textbf{The audit across engines}, random probes, $k=8$: the analytic model as deployed, with the MuJoCo-fitted offset (not refitted), and with the boundary scanned in that engine.}
\label{tab:engines}
\centering\scriptsize
\setlength{\tabcolsep}{1.6pt}
\begin{tabular}{@{}lcccccc@{}}
\toprule
 & \multicolumn{3}{c}{PyBullet} & \multicolumn{3}{c}{PhysX} \\
\cmidrule(lr){2-4}\cmidrule(lr){5-7}
 & deployed & MJ offset & own scan & deployed & MJ offset & own scan \\
\midrule
support escape & $0.167$ & $0.023$ & $0.031$ & $0.039$ & $0.065$ & $0.000$ \\
episode coverage & $0.763$ & $0.951$ & $0.948$ & $0.914$ & $0.872$ & $0.995$ \\
optimism $e_8$ & $+0.232$ & $+0.002$ & $-0.002$ & $+0.369$ & $+0.419$ & $+0.338$ \\
failure & $0.307$ & $0.109$ & $0.107$ & $0.474$ & $0.534$ & $0.469$ \\
\bottomrule
\end{tabular}
\end{table}

\subsection{Physical-arm run}
\begin{figure}[t]
\centering
\includegraphics[width=0.98\columnwidth]{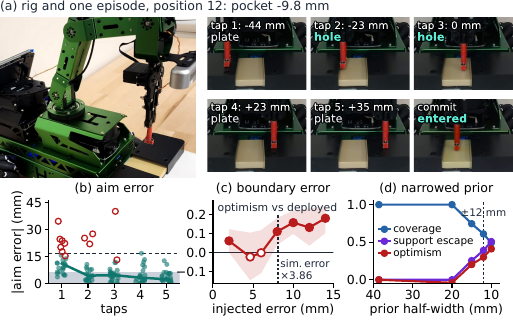}
\vspace{-3pt}
\caption{\textbf{The run on the physical arm and its replay.} (a) The arm at the rig, and one five-tap episode from the front camera, pocket at $-9.8$\,mm: hole read at $-23$ and $0$\,mm, plate elsewhere; the commit at $-11.6$\,mm enters. (b) Aim error against taps, medians joined; filled points entered, open did not; band: the model's window; dashed: the opening's half-width. (c) The recorded taps under an injected boundary error: optimism of the faulted model minus the deployed one, paired over positions, $95\%$ band, filled where resolved; dotted: the simulation's $2.1$\,mm error at rig scale. (d) The same taps under a narrowed prior; at $\pm12$\,mm both conditions fire.}
\label{fig:hardware}
\end{figure}

The insertion pipeline ran on a Hiwonder JetRover (5-DOF arm, Fig.~\ref{fig:hardware}a) inserting a $5.9$\,mm tool into a $34$\,mm pocket machined in a rigid block, the simulation's geometry at $3.86\times$, which preserves its prior-to-window ratio of $5.9$ exactly by construction. The block was slid by hand to positions unknown to the robot. A tap descended in $0.5$\,mm steps until a touch switch on the tool closed, on the block or, past its allowance, through the opening. The robot ran the same particle filter ($2048$ particles, $3\%$ of likelihood mass kept on a misread) and scenario-minimax commit, and the commit was a physical insertion of the same tool. A touch sweep afterwards bracketed the opening for the truth; $12$ of $14$ positions pass the stated guards.

Over $118$ episodes at one to five taps the truth stayed in the retained set on every one, the tool entered on $19$ of $26$ one-tap commits and $24$ of $24$ five-tap commits, and mean aim error fell from $12.6$ to $3.6$\,mm, both resolved by a cluster bootstrap over positions (Table~\ref{tab:hardware}, Fig.~\ref{fig:hardware}b). Predicted failure tracked realized failure at every budget (optimism within $0.12$ of zero, every interval covering zero).

\noindent\textbf{Fault injection on recorded taps.} Taps came from a fixed grid and the frozen schedule, never from the belief, so the $350$ recorded readings replay under any observation model. Replayed as deployed they reproduce the run (coverage $1.000$, optimism $0.000$). With the tap boundary pulled inward, the fault the audit found in simulation, the optimism condition resolves the error from $8$\,mm upward ($+0.11\,[+0.04,+0.18]$ at $8$\,mm, paired over positions; Fig.~\ref{fig:hardware}c), and $8$\,mm is the simulation's $2.1$\,mm error at rig scale. Coverage holds to $14$\,mm under the $\pm38.6$\,mm prior, which is wide against a $34$\,mm opening. With the prior narrowed instead, the truncated-prior fault of Table~\ref{tab:faults}, both conditions fire together: at $\pm12$\,mm coverage falls to $0.61$ with support escape $0.39$ and optimism $+0.30$ (Fig.~\ref{fig:hardware}d), the signature Section~\ref{sec:analysis} reports in simulation, now on recorded taps. Two earlier runs on a paper-sheet rig, one by camera, gave the same picture and ship with the release. The physical-robot demonstrations are also shown in the supplementary video.

\begin{table}[h]
\vspace{5pt}
\caption{\textbf{The run on the physical arm}, by number of taps. Realized failure is the aim outside the model's $6.57$\,mm window of the measured centre; optimism is realized minus predicted. Coverage was $118$ of $118$.}
\label{tab:hardware}
\centering\scriptsize
\setlength{\tabcolsep}{3pt}
\begin{tabular}{@{}lcccccc@{}}
\toprule
taps & episodes & in window & entered & predicted & realized & optimism \\
\midrule
1 & $26$ & $6$ & $19$ & $0.657$ & $0.769$ & $+0.112$ \\
2 & $21$ & $13$ & $18$ & $0.474$ & $0.381$ & $-0.093$ \\
3 & $26$ & $19$ & $24$ & $0.315$ & $0.269$ & $-0.046$ \\
4 & $21$ & $15$ & $21$ & $0.225$ & $0.286$ & $+0.061$ \\
5 & $24$ & $22$ & $24$ & $0.084$ & $0.083$ & $-0.001$ \\
\bottomrule
\end{tabular}
\end{table}

\section{Analysis}\label{sec:analysis}
\begin{figure}[t]
\centering
\includegraphics[width=0.98\columnwidth]{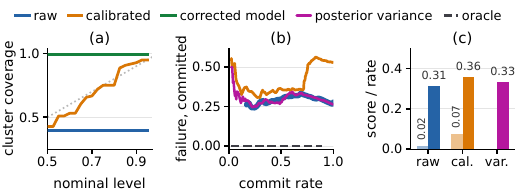}
\vspace{-4pt}
\caption{\textbf{Commit-decision audit on Split~B.} (a) Cluster coverage versus nominal level under random probing ($k=8$). (b) Failure among committed episodes versus commit rate: cover schedule, budgets $1$--$5$, budget-simultaneous calibration, and first-passage commitment. Curves compare scoring and execution jointly; areas are point estimates over each method's attainable range. The non-deployable hindsight oracle achieves zero failure at $89\%$ commitment. (c) Under the same cover schedule at $60\%$ commitment, mean predicted score (pale) versus realized failure (solid); posterior variance makes no probability claim.}
\label{fig:decision}
\end{figure}

\noindent\textbf{The commit decision.} The decision layer uses the budget-simultaneous construction of Section~\ref{sec:calib} and the validation-tuned margin. Probe trajectories are nested, so stopping at budget $k$ uses the trajectory's first $k$ taps; calibration and test are Split~B. The shared threshold is $\widehat t=1.004$ for the cover schedule and $1.018$ for random probes. Three deployable signals commit at the first budget whose value is at most $\tau$: the raw scenario fraction at $\nu=0.8$, the adjusted failure score, and posterior variance; a hindsight oracle on realized outcomes is the non-deployable reference. Sweeping $\tau$ traces failure among committed episodes against commit rate (Fig.~\ref{fig:decision}b). On the cover schedule, the measured areas under these curves are $0.293$ for the raw signal, $0.306$ for posterior variance, and $0.407$ for the adjusted signal; lower is better; under random probing posterior variance is lowest, $0.325$. A monotone transformation of a fixed score leaves a threshold sweep unchanged; our set and margin adjustments change the execution rule too, so these curves compare combined scoring and execution procedures, not recalibration alone.

The evaluated signals cannot rank these failures because the failures are escapes. Call an episode escaped when the truth lies outside the full particle hull at the schedule's final budget $k=5$; claimed score and realized failure are taken at each committed episode's own stopping budget. At the threshold committing $60\%$ on the cover schedule, the committed pool holds $43$ escaped episodes of $230$ against $7$ of the $154$ rejected for the raw signal, $41$ of $231$ against $9$ of $153$ for the adjusted signal, and $43$ of $230$ against $7$ for posterior variance. Selection re-creates the optimism the repair removed: at commit the raw signal claims $0.017$ and realizes $0.313$, the adjusted signal $0.075$ and $0.359$ (Fig.~\ref{fig:decision}c). The oracle reaches zero failure at an $89\%$ commit rate and no evaluated signal closes the gap: a gate on belief confidence retains confidently wrong instances at four times the rate of the pool it rejects. Nor does the data-belief channel see them. Prequential surprise, the negative log predictive probability of each tap before it is assimilated, aggregated by the maximum over the taps seen, detects final-budget escape on the cover schedule ($50$ escaped, $334$ covered test episodes) with rank AUC $0.85$ at $k=1$ but $0.54$ at $k=5$, cluster-bootstrap interval $[0.46,0.63]$. The escapes that persist are wrong hypotheses that \emph{explain} their taps.

\noindent\textbf{Fault classes and their signatures.} Does the audit point to the class of failure? Table~\ref{tab:faults} injects five controlled mismatch classes one at a time on top of the corrected model and records what the audit reports and which remedy resolves each: noise, simulator tap noise at $4\times$ the modeled standard deviation, fixed by scaling the filter's $\sigma$ by four; boundary, the original boundary of \eqref{eq:tap}, fixed by the corrected one; prior, truths drawn from $\pm12$\,mm under a $\pm10$\,mm prior, fixed by widening the prior to $\pm12$\,mm; particles, $128$ without jitter, fixed by $1024$ with jitter; frame, $1$\,mm added to the committed aim on the simulator side, fixed by removing it.

\begin{figure}[t]
\centering
\includegraphics[width=0.98\columnwidth]{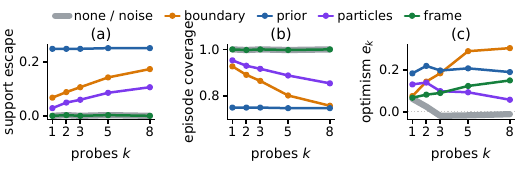}
\vspace{-2pt}
\caption{\textbf{The five classes over the probe budget}, random probes, corrected model, $n=384$ per point. Zero-mean noise matches no fault at every budget, so they share a line; both, and the frame offset, escape on at most one episode in $384$.}
\label{fig:signatures}
\end{figure}

The classes produce different combinations of escape, coverage, optimism and surprise; the table is not a fault classifier. The boundary error and the truncated prior look alike at $k=8$, both failing coverage with optimism, but read over the budget they separate (Fig.~\ref{fig:signatures}): the prior escapes on $0.25$ of episodes from the first tap and fails all four repeats at once, while the boundary error climbs from $0.07$ to $0.17$ as evidence arrives and optimism climbs with it, from $+0.07$ to $+0.30$. Particle impoverishment raises support escape while the score residual \emph{improves}, and the frame offset is invisible to the belief at every budget (coverage $1.00$) while the executor's optimism grows and it fails $0.26$: the signature of a latent the model lacks. Tempering, $p^{1/T}$ on the likelihood so that $T=4$ doubles the effective $\sigma$, resolves none of them: a $30\sigma$ residual is still $15$ effective standard deviations; development-tuned alternatives follow below. Paired over truth clusters, the known-fault corrections change failure by $-0.281\,[-0.352,-0.214]$ for the boundary error, $-0.161\,[-0.219,-0.107]$ for the frame offset, and $-0.104\,[-0.167,-0.044]$ for the truncated prior. The prior correction widens the prior to $\pm12$\,mm, so its $0.18$ failure rate is measured on the wider task, not evidence of an uncorrected prior fault. For particle impoverishment, the paired change is $-0.026\,[-0.057,+0.003]$, which does not establish a failure reduction.

\begin{table}[h]
\vspace{5pt}
\caption{\textbf{Controlled mismatch tests.} Random probes, $k=8$, Split~B. Fig.~\ref{fig:signatures} shows the corresponding audit metrics. Noise at $4\times$ has no observed effect, the two readings being $30\sigma$ apart. Surprise is the prequential-surprise AUC for escape; other numeric columns report failure before and after tempering ($T=4$) or known-fault correction. Cluster-bootstrap half-widths reach $0.077$; paired differences are reported in the text.}
\label{tab:faults}
\centering\scriptsize
\setlength{\tabcolsep}{4pt}
\begin{tabular}{@{}lcccc@{}}
\toprule
fault & fail & surprise & temp.\ $T{=}4$ & known fix \\
\midrule
none (baseline) & $0.10$ & -- & -- & -- \\
noise & $0.10$ & -- & $0.10$ & $0.10$ \\
boundary & $0.38$ & $0.64$ & $0.38$ & $0.10$ \\
prior & $0.29$ & $0.59$ & $0.29$ & $0.18$ \\
particles & $0.13$ & $0.67$ & $0.13$ & $0.10$ \\
frame & $0.26$ & -- & $0.26$ & $0.10$ \\
\bottomrule
\end{tabular}
\end{table}

\noindent\textbf{Robust filters as an alternative.} Three alternatives were tuned on the $24$ development instances under the model as deployed (lowest $k=8$ failure, ties to the first grid value) and then confronted with the boundary error: tempering, $T\in\{1,2,4,8,16\}$, chosen $T=1$; a contamination mixture $(1-\epsilon)p+\epsilon q_0$ with $q_0$ uniform over the $19$\,mm observable height range, $\epsilon\in\{0.01,0.05,0.1,0.2\}$, chosen $0.01$; and rejuvenation, redrawing a fraction $\{0.02,0.05,0.1\}$ of the particles from the prior at each resampling, chosen $0.02$. The mixture cuts support escape from $0.174$ to $0.042$ and rejuvenation to $0.062$, yet neither gives a statistically resolved reduction in failure, $+0.016\,[-0.008,+0.039]$ and $+0.029\,[-0.003,+0.062]$ paired, with optimism still above $+0.29$. When the model is right the mixture costs failure, $0.102$ to $0.167$. A tuned robust filter repairs support, not the decision; of the alternatives tested here, only the known-fault correction repairs both conditions.

\section{Limitations and Conclusion}
\noindent\textbf{Limitations.} The task evaluations are in simulation; on hardware the tool is slender against the modelled peg, the truth shares the tap's sensor, and the prior is wide against the opening, so coverage there is the weaker test. The protocol assumes the task state can be restored between probes, as in a fixtured cell; the cross-engine experiments test portability across solvers, not the reality gap. The conformal guarantee is marginal coverage of the decision-coordinate interval under exchangeability, and certifies neither the failure score nor risk conditional on commitment. The fitted correction transfers to PyBullet, not to PhysX, and the audit's signatures guide investigation rather than naming a cause.

\noindent\textbf{Conclusion.} Auditing decision-coordinate coverage and mean executor optimism separately reveals distinct failures of a probe-then-commit robot. Calibration restores coverage while the evaluated gates still select confidently wrong episodes; the audit's signatures led to a model error whose frozen correction held on new instances and transferred to one engine of two, the audit saying which. Calibration and inspection are needed together; on a physical arm the pipeline inserts a tool into a rigid pocket, and its recorded taps show the audit's signatures under injected faults.

\section*{Acknowledgment}
Claude was used to assist with development and debugging of experimental software and data processing. All AI-assisted content was reviewed and validated by the authors.

{\footnotesize
\bibliographystyle{IEEEtranN}
\bibliography{refs}}

\end{document}